\documentclass[conference]{IEEEtran}
\IEEEoverridecommandlockouts
\usepackage{cite}
\usepackage{amsmath,amssymb,amsfonts}
\usepackage{graphicx}
\usepackage{textcomp}
\usepackage{xcolor}
\usepackage{bm}
\usepackage{amssymb}
\usepackage{amsthm}
\usepackage{multirow}
\usepackage{booktabs}
\usepackage{algorithmicx,algorithm}
\usepackage{stfloats}
\usepackage{framed}
\usepackage{subfigure}
\usepackage{bbding}
\usepackage{fancyhdr}

\def\BibTeX{{\rm B\kern-.05em{\sc i\kern-.025em b}\kern-.08em
    T\kern-.1667em\lower.7ex\hbox{E}\kern-.125emX}}
\begin{document}
\bibliographystyle{unsrt}
\title{FunduSAM: A Specialized Deep Learning Model for Enhanced Optic Disc and Cup Segmentation in Fundus Images
\thanks{*: Corresponding Author.}
}
\author{\IEEEauthorblockN{1\textsuperscript{st} Jinchen Yu}
\IEEEauthorblockA{\textit{University of Science and Technology of China} \\
Anhui, China \\
jxgfyjc@mail.ustc.edu.cn}
\and
\IEEEauthorblockN{2\textsuperscript{nd} Yongwei Nie}
\IEEEauthorblockA{\textit{South China University of Technology} \\
Guangdong, China \\
nieyongwei@scut.edu.cn}
\and
\IEEEauthorblockN{3\textsuperscript{rd} Fei Qi}
\IEEEauthorblockA{\textit{South China University of Technology } \\
\textit{Guizhou Minzu University}\\
Guangdong, China \\
fqiscut@foxmail.com}
\and
\IEEEauthorblockN{4\textsuperscript{th} Wenxiong Liao}
\IEEEauthorblockA{\textit{South China University of Technology} \\
Guangdong, China \\
cswxliao@mail.scut.edu.cn}
\and
\IEEEauthorblockN{5\textsuperscript{th} Hongmin Cai$^*$}
\IEEEauthorblockA{\textit{South China University of Technology} \\
Guangdong, China \\
hmcai@scut.edu.cn}
}
\maketitle
\thispagestyle{fancy}
\fancyhead{}

\cfoot{\quad}
\renewcommand{\headrulewidth}{0pt}
 \renewcommand{\footrulewidth}{0pt}

\begin{abstract}
The Segment Anything Model (SAM) has gained popularity as a versatile image segmentation method, thanks to its strong generalization capabilities across various domains. However, when applied to optic disc (OD) and optic cup (OC) segmentation tasks, SAM encounters challenges due to the complex structures, low contrast, and blurred boundaries typical of fundus images, leading to suboptimal performance. To overcome these challenges, we introduce a novel model, FunduSAM, which incorporates several Adapters into SAM to create a deep network specifically designed for OD and OC segmentation. The FunduSAM utilizes Adapter into each transformer block after encoder for parameter fine-tuning (PEFT). It enhances SAM’s feature extraction capabilities by designing a Convolutional Block Attention Module (CBAM), addressing issues related to blurred boundaries and low contrast. Given the unique requirements of OD and OC segmentation, polar transformation is used to convert the original fundus OD images into a format better suited for training and evaluating FunduSAM. A joint loss is used to achieve structure preservation between the OD and OC, while accurate segmentation. Extensive experiments on the REFUGE dataset, comprising 1,200 fundus images, demonstrate the superior performance of FunduSAM compared to five mainstream approaches.
\end{abstract}


\begin{figure*}[!t]
    \centering
    \includegraphics[width = 16cm]{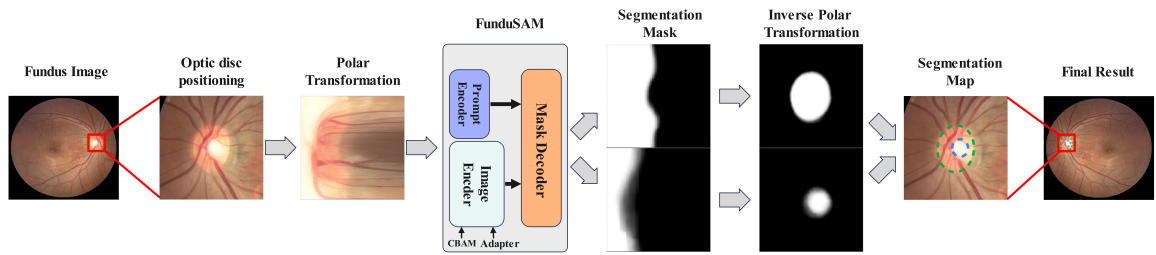}
\caption{The pipeline of our method for optic disc and cup segmentation, the input fundus image is first transformed in polar coordinates, followed by learning and prediction of the image using our proposed FunduSAM, and finally the output is transformed in inverse polar coordinates to obtain the result}
    \label{fig1_pipeline}
\end{figure*}

\section{Introduction}

Glaucoma, a chronic eye disease, can lead to irreversible vision loss and is currently the second leading cause of blindness worldwide, surpassed only by cataracts. Due to its irreversibility at onset, early screening and accurate diagnosis are crucial for glaucoma patients. 
Clinically, the Cup-to-Disc Ratio (CDR) \cite{WOS:000087387300023} is commonly used to assess the condition by calculating the ratio of vertical cup diameter (VCD) to vertical disc diameter (VDD). However, manual assessment of CDR measurements is subjective, time-consuming, and resource-intensive.

In the field of medical imaging, deep learning-based artificial intelligence techniques have achieved significant success. 
Fully Convolutional Networks (FCN) \cite{long2015fully} is the first to apply fully convolutional layers to image segmentation tasks, allowing the network to accept an input image of arbitrary size and output a segmentation map of the same size as the input. 
The subsequently proposed U-Net \cite{WOS:000365963800028} , with its encoder-decoder architecture, has achieved remarkable progress in medical image segmentation.
DeepLabv3+ \cite{WOS:000426687100005} further introduced encoder-decoder architecture along with dilated convolutions and the Atrous Spatial Pyramid Pooling (ASPP) module, enhancing segmentation accuracy. 
Mask R-CNN \cite{WOS:000425498403005} adds a branch to Faster R-CNN, which performs well in a variety of tasks.
Recently, the Segment Anything Model (SAM) \cite{WOS:001159644304024}, based on a U-Net-like encoder-decoder structure, has demonstrated excellent robustness and interactivity in the field of image segmentation, as well as good accuracy and generalization ability in a variety of tasks. This interactivity is especially critical in the medical field, allowing for more effective differentiation of the various complex architectures of medicine when prompted by human annotations, especially in the case of fundus images.
However, SAM’s performance in medical image segmentation is limited by the lack of guidance from domain-specific medical knowledge in pre-trained models. Additionally, medical images often present challenges such as low contrast, blurred tissue boundaries, and small lesion regions, which further constrain SAM’s performance in medical image segmentation.

To address this issue, this paper proposes a novel model called FunduSAM, which introduces various adaptive strategies based on the SAM framework to construct a deep network better suited for fundus image segmentation.
The pipeline of our method as shown in Fig.\ref{fig1_pipeline}. Specifically the contributions of our work are summarized as follows:

1) We propose FunduSAM for the fundus optic disc (OD) and optic cup (OC) segmentation task. Leveraging an Adapter strategy built upon the SAM to achieve parameter efficient finetuning (PEFT), FunduSAM adapts more effectively to the specific fundus image segmentation task. Additionally, the inclusion of the Convolutional Block Attention Module (CBAM) further enhances feature extraction capabilities from fundus images.

2) FunduSAM designs a fundus images preprocessing strategy using polar transformation. This manner balances the proportions of different regions, simplifies boundary information, and leverages the prior knowledge of OD and OC boundaries to guide the segmentation process, resulting in improved segmentation performance.

3) In the fundus image, for the structural priori knowledge between OD and OC, we consider the use of a joint loss function. The joint loss function of FunduSAM combines cross-entropy losses for individual OD and OC segmentation tasks. For the purpose of enhancing segmentation accuracy, a structural constraint loss function is also incorporated.

4) The effectiveness of the proposed FunduSAM is evaluated through comparison experiment and ablation study on the REFUGE dataset. The results demonstrate that FunduSAM achieves favorable segmentation performance.

\section{Related Work}

\subsection{SAM-based Applications in the Medical Field}
Due to the powerful segmentation performance and generalization ability of the SAM, many works have applied it to image segmentation tasks in different domains. However, the performance of SAM in medical domain is more limited due to its lack of medical specific knowledge and facing challenges such as low image contrast, fuzzy tissue boundaries and tiny lesion regions in medical images. And in order to solve these challenges and apply SAM to the medical field, some methods have been proposed nowadays, SAMed \cite{zhang2023customizedsegmentmodelmedical} utilizes a low-rank (Lora) strategy to fine-tune the local parameters of SAM, which reduces the computational overhead and makes it more adaptable to the medical field. MedSAM \cite{WOS:001148371500004}, through a large amount of data and freezing the image encoder and the prompt Encoder, the Mask Decoder was completely fine-tuned. AutoSAM \cite{INSPEC:23394927} changed the original prompter encoder structure of SAM and explored three unprompted prediction headers. Med-SA \cite{wu2023medicalsamadapteradapting} configured adapter correlation network layers for each of the three parts that make up the SAM and only fine-tuned it locally. SAMUS \cite{INSPEC:23780196} introduced a CNN branch and also used cross-branching attention to improve the performance of SAM in the medical domain.

\subsection{Parameter Effective Fine Tuning (PEFT) }
Parameter Efficient Fine-Tuning (PEFT) is a strategy for efficiently fine-tuning models by selectively fine-tuning a small fraction (typically less than 5\% of the total parameters) of the model parameters while keeping the majority of the parameters frozen, which avoids catastrophic forgetting and minimizes computational and storage requirements. PEFT includes adapter-based techniques, selective parameter tuning, prompt-driven fine-tuning, and Low Rank Adaptation (LoRA). In the adapter paradigm \cite{WOS:000684034302095}, small layers of trainable parameters are added to each layer of a pre-trained model, enabling the model to adapt to a new task without significantly increasing the number of parameters. Some approaches \cite{guo2021parameterefficienttransferlearningdiff} proposed a selective fine-tuning strategy, which reduces the cost of fine-tuning by updating the parameters of only a few layers, while maintaining high task performance. Prompt tuning \cite{WOS:000855966303015} \cite{WOS:000698679200153} proposes tuning the model's behavior by using cues as additional inputs or inserting cues in the middle layer of the model, respectively. LoRA \cite{hu2021lora} introduced trainable low-rank matrices in the transformer layer for weight updating.

The PEFT technique has also been shown to be effective in the field of computer vision (CV). Recent studies have shown that adaptive techniques can be easily applied to a variety of downstream computer vision tasks \cite{article} \cite{chen2022adaptformeradaptingvisiontransformers}. Therefore, we believe that the adaptive approach is the best choice for introducing SAM into the medical field. We foresee that this concise and powerful method will bring broader possibilities for the development of basic medical models.

\begin{figure*}[!t]
    \centering
    \includegraphics[width = 16cm]{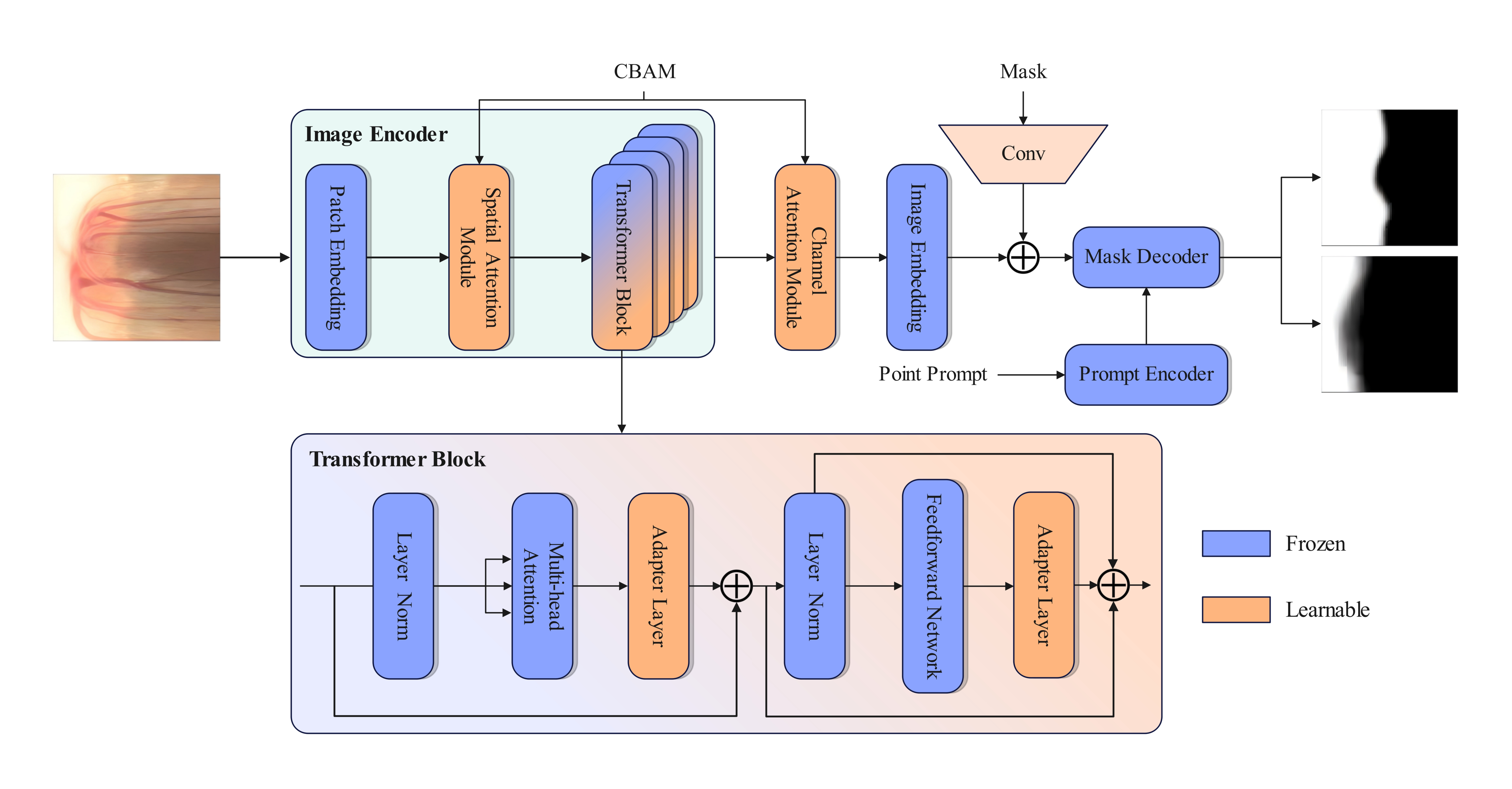}
\caption{Overview of the proposed FunduSAM consists of Adapter layers, Spatial Attention Module and Channel Attention Module in CBAM}
    \label{fig2_framework}
\end{figure*}

\section{Methods}
We propose FunduSAM, a novel deep-learning framework designed specifically for OD and OC segmentation tasks. As the overall architecture of FunduSAM is adjusted based on SAM, we have not modified the structure and parameters of the prompt encoder and mask decode, to retain SAM's advantages in processing prompt information and generating segmentation masks. However, we have made several changes to the image encoder. The framework of the model is shown in Fig.\ref{fig2_framework}. 

Specifically, based on the specificity of fundus image data, the model utilizes polar transformation to balance the OC ratio, simplifies the boundary information, and introduces structural priori knowledge.
Then, preprocessed fundus images are encoded into image embeddings by an image encoder based on the standard Visual Transformer (ViT) feature map. The image encoder comprises sixteen Transformer blocks, including twelve window attention blocks and four global attention blocks. The window attention blocks handle local information using window-based partition attention mechanisms, while the global Attention blocks capture the overall context of the image.
To fine-tune the image encoder, two adapter layers are configured for each transformer block, further performing localized PEFT on the network. Additionally, we enhance feature extraction capabilities for blurry boundaries and low-contrast medical images by disassembling a CBAM and put it into the front and back ends of the image encoder.
We simultaneously incorporate cross-entropy losses for both OD and OC, and introduce a structural constraint loss function to create the comprehensive total loss for the segmentation task.

\subsection{Adapters in the Vit Block}
Image Encoder is the part of the SAM network with the largest number of parameters, and its complete fine-tuning would bring great operational burden and computational cost. Therefore, the introduction of adapter technology by the PEFT is an effective means to help us freeze the parameters in the original model Image Encoder and iteratively update only the parameters in the adapter module. In our implementation, we deploy two adapters for each Transformer block, the first one is placed after the multi-head attention mechanism, and the second one is placed after the feedforward layer. Both adapters have the same components including down projection, activation function and sequential up projection. From there, the transformer block is locally fine-tuned to make the whole model more adapted to the task of fundus OD and OC segmentation.

\subsection{CBAM}
The Convolutional Block Attention Module (CBAM) \cite{WOS:000594221500001} consists of a spatial attention module and a channel attention module. Inspired by the work of \cite{WOS:000845070100011}, we disassemble the CBAM and put it into the front end and back end of the Image Encoder, which consists of 16 Transformer Blocks, respectively. The function of the spatial attention module is to model the spatial information of the feature maps before the feature maps are entered into image encoder, which in turn leads to better contextual modeling of the feature map information in the transformer block. The function of the channel attention module is to dynamically adjust the weight of each channel according to its importance, and feature assignment can be performed for different tasks. As a result, changes in interest of feature mapping can be better recognized. Utilizing CBAM to introduce two attention mechanisms at the same time, for the polar coordinate transformed fundus image, it can make both the image processing and feature extraction parts of the image encoder more thorough, thus obtaining a more accurate segmentation effect.

\subsection{Polar Transformation}
\begin{figure}[!h]
    \centering
    \includegraphics[width = 7.5cm]{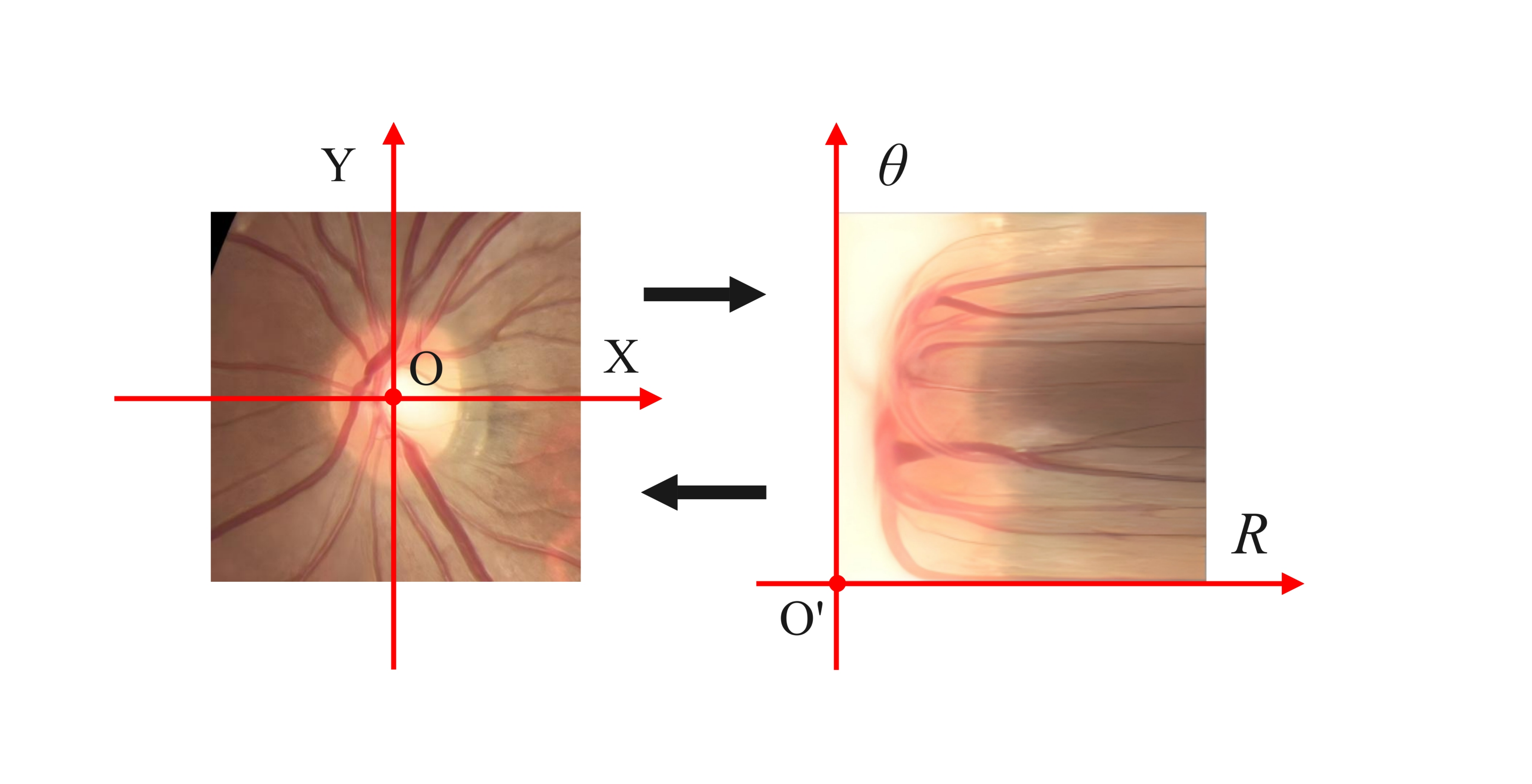}
\caption{Visualization of polar transformation, which simplifies the boundary information and balances the proportion of OD and OC.}
    \label{fig4_polar}
\end{figure}

We designed a polar coordinate mapping strategy for preprocessing fundus images. Specifically, we used the method from Wang et al. [18] to crop the OD, and the cropped image is converted from a Cartesian coordinate system image to a polar coordinate system image. Similarly, we apply an inverse polar transformation to the final output mask. Fig.\ref{fig4_polar} shows a comparison of fundus image before and after polar transformation.

Take the center of the ROI region as the origin, $r$ is the distance from a point to the origin, and $\theta$ is the angle. Then any point $P(x, y)$ can be converted to a point $P'(\theta, r)$ in the polar coordinate system. The conversion relation from the Cartesian coordinate system to the polar coordinate system is
\begin{equation}
        \begin{aligned}
          \begin{cases}
            x =r\cos \theta \\
            y= r\sin \theta
            , 
            \end{cases}
        \end{aligned}
      \end{equation}

while the conversion relation from the polar coordinate system to the Cartesian coordinate system is

\begin{equation}
        \begin{aligned}
          \begin{cases}
            r=\sqrt{x^2+y^2} \\
            \theta=\tan^{-1} \frac{y}{x}
            .
            \end{cases}\\
        \end{aligned}
      \end{equation}
      
Since the proportion of OC in the original image is extremely uneven, which easily leads to bias and overfitting in training segmentation network, we use polar transformation to balance the distribution proportion of OD and OC \cite{Fu_2018}. And at the same time, it makes the original ellipsoidal boundary become simpler and easy to segment, which is conducive to CBAM extraction of image features, and thus promote the training and performance of the network.

\subsection{Loss Function}
Since there exists a structural a priori knowledge that the OC must be contained in the OD in the task, we are inspired by the work of \cite{WOS:000371029500008}, to consider using a joint loss function, which is a model that learns this a priori knowledge while segmenting the two and enhances each other to improve the segmentation accuracies of both OD and OC together. Our joint loss function is defined as
\begin{equation}
        \begin{aligned}
          L=\omega _1L_{disk}+\omega _2L_{cup}+\omega _3L_{contain}, 
        \end{aligned}
      \end{equation}
where $\omega _1, \omega _2, \omega _3$ are the weight parameters and update within the range of $[0,1]$, $L_{disk}$ and $L_{cup}$ are the cross-entropy loss functions for segmenting the OD and OC, respectively. $L_{contain}$ introduces a structural constraint between the OD and OC. The definition of $L_{disk}$ and $L_{cup}$ are as follows:
\begin{equation}
        \begin{aligned}
          L_{disk}=-\frac{1}{N} \sum_{i=1}^{N}\left[x_{i} \log \left(\hat{x}_{i}\right)+\left(1-x_{i}\right) \log \left(1-\hat{x}_{i}\right)\right], 
        \end{aligned}
      \end{equation}

\begin{equation}
        \begin{aligned}
          L_{cup}=-\frac{1}{N} \sum_{i=1}^{N}\left[y_{i} \log \left(\hat{y}_{i}\right)+\left(1-y_{i}\right) \log \left(1-\hat{y}_{i}\right)\right], 
        \end{aligned}
      \end{equation}

$L_{contain}$ is defined as

\begin{equation}
        \begin{aligned}
          L_{contain}= { \sum_{i=1}^{N}} (y_i\cdot(1-x_i)),    
        \end{aligned}
      \end{equation}
where $N$ is the total number of pixels in the image, and $x_i$, $y_i$ are the binary labels (0 or 1) in the OD mask and the OC mask respectively, $\hat{x}_{i}$, $\hat{y}_{i}$ are the predicted probability of the $i$-th sample belonging to class 1 correspondingly.

\section{Experiment and Results}
In this section, we evaluate the performance of our proposed method by comparing it with five state-of-the-art (SOTA) methods on REFUGE dataset and conducting ablation experiment.

\subsection{Datasets}
For segmentation of fundus images, we conducted experiments using the REFUGE \cite{fang2022refuge2challengetreasuretrove} dataset. The REFUGE glaucoma competition dataset consists of a total of 1,200 images, which are divided into three parts, namely the training set, the validation set, and the test set. Each of them has 400 images. The training set has an image size of 2,124 × 2,056 pixels, and the validation and test set images were 1,634 × 1,634 pixels. Due to the different shooting cameras, there are large differences in the color, texture and contrast. Between the training set and validation and test set images, it exists the domain adaptation problem. But this problem is solved by centering, cropping and other alignment operations. In this paper, we randomly divide 1200 images into a training set of 960 and a test set of 240 in the ratio of 8:2 to verify the effectiveness of FunduSAM.

\subsection{Implementation Details}
The experiment uses PyTorch deep learning framework and OpenCV image processing library, Ubuntu 24.04 operating system, NVIDIA A100-PCIE-40GB graphics card, Adam optimizer, with an initial learning rate of 0.0001. Batchsize training phase set to 32, test phase set to 1 and a total of 150 epochs of training. To simplify the experimental process, we randomly sample a point in the foreground region of the label to simulate the prompt process. The loss function is the joint loss of OD and OC cross entropy and structural constraints a priori, and the evaluation metrics include intersection-over-union (IOU) and Dice coefficient (Dice)\cite{milletari2016vnetfullyconvolutionalneural}.

\subsection{Comparison Experiment}
In order to validate the general performance of our proposed model, we compared other SOTA segmentation methods. The quantitative results are shown in the Table \ref{tab:comparison}. In the table, we compared FunduSAM with recognized medical image segmentation methods, including ResUnet \cite{Zhang_2018}, nnUNet \cite{isensee2021nnu}, TransUNet \cite{chen2021transunettransformersmakestrong}, Swin-UNetr \cite{hatamizadeh2022swinunetrswintransformers}, Fully fine-tuned SAM (MedSAM)\cite{WOS:000471732700007}.

\begin{table}[!h]
\caption{Quantitative comparison of our FunduSAM and other methods on REFUGE data. The performance is evaluated by the Dice score and IOU}
\begin{center}
      \scalebox{1.2}{
        \begin{tabular}{ccc|cc}
        \hline \multirow{2}{*}{ Method } & \multicolumn{2}{c|}{ Optic-Disc } & \multicolumn{2}{c}{ Optic-Cup } \\
        \cline { 2 - 5 } & Dice & IOU & Dice & IOU \\
        \hline ResUNet & 0.929 & 0.855 & 0.801 & 0.723 \\
        \hline nnUNet & 0.947 & 0.873 & 0.849 & 0.751 \\
        \hline TransUNet & 0.950 & 0.877 & 0.856 & 0.759 \\
        \hline Swin-UNetr & 0.953 & 0.879 & 0.843 & 0.745 \\
        \hline MedSAM & 0.946 & 0.867 & 0.828 & 0.759 \\
        \hline FunduSAM & \textbf{0.961} & \textbf{0.882} & \textbf{0.867} & \textbf{0.789}\\
        \hline
        \end{tabular}}
      \end{center}
      \label{tab:comparison}
    \end{table}
    
The above table demonstrates the performance and evaluation of the five methods compared to our model in the task of fundus OD and OC segmentation, and our method achieves the highest results for both Dice and Iou in the task. Taking the Dice coefficient of OD and OC as a reference, our method improves 3.44\%, 8.24\% over ResUNet, 1.48\%, 2.12\% over nnUNet, 1.16\%, 1.29\% over TransUNet, 0.84\%, 2.85\% over Swin-UNetr, and 1.59\%, 4.71\% over MedSAM, which proves the superiority of our method.

\begin{figure*}[!b]
    \centering
    \includegraphics[width = 15cm]{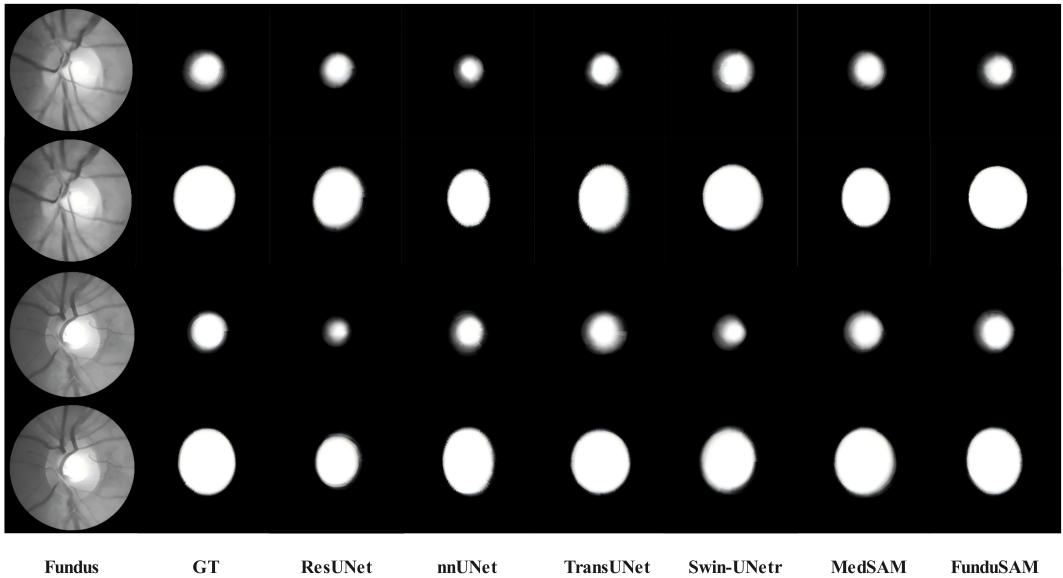}
\caption{Visualization of the results of different methods for fundus optic disc and cup segmentation}
    \label{fig3_visualization}
\end{figure*}

\subsection{Ablation Study}

In order to verify the effectiveness of the three modules we mentioned in this paper, CBAM, adapter and polar transformation, we conducted ablation study on them by selecting the network to which all three modules were added as a control and removing any one of the three modules, and conducting comparison experiments. 

\begin{table}[!h]
\caption{Ablation study on different component combinations of FunduSAM on the optic disc and cup segmentation, PT represents the polar transformation}
\begin{center}
\scalebox{1.0}{
\begin{tabular}{cccc|cc|cc} 
\multirow{2}{*}{SAM} & \multirow{2}{*}{ CBAM } & \multirow{2}{*}{ adapter } & \multirow{2}{*}{PT} & \multicolumn{2}{|c|}{ Optic-Disc } & \multicolumn{2}{c}{ Optic-Cup } \\
& & & & Dice & IOU & Dice & IOU \\
\hline \Checkmark  &  \XSolidBrush  &  \XSolidBrush  &  \XSolidBrush  & 0.829 & 0.755 & 0.721 & 0.638 \\
 \Checkmark  &  \Checkmark  &  \Checkmark  &  \XSolidBrush  & 0.906 & 0.816 & 0.832 & 0.751 \\
 \Checkmark  &  \Checkmark  &  \XSolidBrush    &  \Checkmark  & 0.893 & 0.826 & 0.813 & 0.721 \\
 \Checkmark  &  \XSolidBrush    &  \Checkmark  &  \Checkmark  & 0.957 & 0.875 & 0.855 & 0.778 \\
 \Checkmark  &  \Checkmark  &  \Checkmark  &  \Checkmark  & \textbf{0.961} & \textbf{0.882} & \textbf{0.867} & \textbf{0.789}\\
\hline
\end{tabular}}
\end{center}
\label{tab:ablation}
\end{table}

As shown in Table \ref{tab:ablation}, it is found that the addition of adapter has the greatest effect on the OD and OC segmentation, followed by polar transformation, and finally CBAM. The best results are obtained when all three modules are added, compared to the lack of Adapter, PT, and CBAM, the enhancement of the Dice of the OD segmentation is 7.61\%, 6.07\%, and 0.42\%, the enhancement of the IOU of the OD segmentation is 8.70\%, 7.38\%, 1.37\%, the enhancement of the Dice of the OC segmentation is 4.33\%,6.77\%, 1.52\%, and the enhancement of the IOU of the OC is 5.06\%, 9.43\%, 1.41\%.

\subsection{Visualization}
In this section, we perform visual tests with high contrast and low contrast to demonstrate the superiority of the segmentation performance of our method. In the segmentation visualization, we split the fundus image with the polar transformation, followed by the inverse polar transformation of the OD and OC segmentation masks. The test results of our FunduSAM and each method are shown in the Fig.\ref{fig3_visualization}, the first two lines are the result of the segmentation of the OC and OD under low contrast, and the latter two lines are the results of high contrast. It can be seen that in the case of high contrast, all methods except ResUNet perform well, while in the case of low contrast, other methods have some prediction biases, and only FunduSAM is still stable, which proves the robustness and accuracy of our proposed method.

\section{Conclusion}
In this paper, we propose FunduSAM, a specialized deep learning model for fundus OC and OD segmentation. Specifically, we introduce an Adapter strategy, CBAM, and polar transformation on the network structure of the SAM, which can achieve enhanced feature extraction and PEFT, as well as simplification and balance of the boundary and proportion. Also, we design a joint loss function for this task, thus introducing a priori knowledge of the fundus structure. Extensive experiments show that FunduSAM is effective and superior to the five state-of-the-art methods, showing excellent performance in fundus OD and OC segmentation task.

\bibliography{reference}
\end{document}